\documentclass{article}

\usepackage{PRIMEarxiv}

\usepackage[utf8]{inputenc} 
\usepackage[T1]{fontenc}    
\usepackage{hyperref}       
\usepackage{url}            
\usepackage{booktabs}       
\usepackage{amsfonts}       
\usepackage{nicefrac}       
\usepackage{microtype}      
\usepackage{lipsum}
\usepackage{fancyhdr}       
\usepackage{graphicx}       
\usepackage{subfiles}
\usepackage{amsmath,amssymb,amsfonts}
\usepackage{graphicx}
\usepackage{subfigure}
\usepackage{multirow} 
\usepackage{color}
\pdfoutput=1
\graphicspath{{media/}}     

\renewcommand{\vec}[1]{\boldsymbol{#1}}
  
\title{scRDiT: Generating single-cell RNA-seq data by diffusion transformers and accelerating sampling
}

\author{
  Shengze Dong, Zhuorui Cui, 
  Ding Liu\thanks{Corresponding author. E-mail: liuding@tiangong.edu.cn }, \\
  School of Computer Science and Technology\\
  Tiangong University \\
  Tianjin, 300387\\
  People's Republic of China\\
  \And
  Jinzhi Lei\thanks{Corresponding author. E-mail: jzlei@tiangong.edu.cn}\\
  School of Mathematical Sciences\\
  Tiangong University \\
  Tianjin, 300387\\
  People's Republic of China\\
}

\begin{document}
\maketitle

\begin{abstract}

\textbf{Motivation: }
Single-cell RNA sequencing (scRNA-seq) is a groundbreaking technology extensively utilized in biological research, facilitating the examination of gene expression at the individual cell level within a given tissue sample. While numerous tools have been developed for scRNA-seq data analysis, the challenge persists in capturing the distinct features of such data and replicating virtual datasets that share analogous statistical properties. 

\textbf{Results: }
Our study introduces a generative approach termed scRNA-seq Diffusion Transformer (scRDiT). This method generates virtual scRNA-seq data by leveraging a real dataset. The method is a neural network constructed based on Denoising Diffusion Probabilistic Models (DDPMs) and Diffusion Transformers (DiTs). This involves subjecting Gaussian noises to the real dataset through iterative noise-adding steps and ultimately restoring the noises to form scRNA-seq samples. This scheme allows us to learn data features from actual scRNA-seq samples during model training. 

Our experiments, conducted on two distinct scRNA-seq datasets, demonstrate superior performance. Additionally, the model sampling process is expedited by incorporating Denoising Diffusion Implicit Models (DDIM). scRDiT presents a unified methodology empowering users to train neural network models with their unique scRNA-seq datasets, enabling the generation of numerous high-quality scRNA-seq samples.

\textbf{Availability and implementation: }
\url{https://github.com/DongShengze/scRDiT}

\end{abstract}


\section{Introduction}

Single-cell RNA sequencing (scRNA-seq) is a powerful technique that allows researchers to analyze the gene expression profiles of individual cells within a tissue or population. By capturing and sequencing the RNA molecules within each cell, scRNA-seq offers a comprehensive snapshot of the cellular composition and molecular heterogeneity of a sample\cite{wang2009rna}. This technology has significantly advanced biological research, facilitating an unprecedented exploration of cellular processes at the single-cell level. scRNA-seq has emerged as an indispensable asset in modern biology, shedding light on crucial aspects such as cellular diversity, gene regulation, and the mechanisms underlying various diseases \cite{marguerat2010rna,shen2014rmats,mccarthy2012differential,peng2012comprehensive}.

Since the advent of the first scRNA-seq study was published in 2009\cite{Tang2009}, a multitude of data analysis methods has emerged, addressing diverse aspects such as cell clustering\cite{Kiselev:2017aa,Stuart:2019aa,Tan:2019aa,Pliner:2019aa}, rare cell type detection\cite{Johansen:2019aa,Tsoucas:2018aa,Jindal:2018aa}, differentially expressed gene indentification\cite{Song:2021aa,Kharchenko:2014aa,Van-den-Berge:2020aa}, and trajectory inference\cite{Trapnell:2014aa,Qiu:2017aa,Street:2018aa,Cao:2019aa}. These methodologies are instrumental in uncovering specific patterns in scRNA-seq data. However, despite their value, a comprehensive understanding of the data features is still in its infancy, leaving researchers grappling with the challenge of formulating scRNA-seq data through simplistic rules. 

A prominent limitation of scRNA-seq data stems from the scarcity of available samples. The cost-intensive nature of single-cell RNA sequencing creates financial barriers, impeding widespread access. Additionally, ethical considerations and the limited availability of patients for specific diseases may curtail the clinical application of this technique. The resulting dearth of scRNA-seq samples compromises the overall depth of information and introduces uncertainty into downstream tasks, casting doubt on the reliability of obtained results\cite{button2013power}.

To address these challenges, a promising avenue involves the development of computational methods to generate in silico datasets. Several simulators, such as scDesigns\cite{WeiVivianLi2019,scDesign2}, ZINB-WaVE\cite{Risso2018}, Splatter\cite{Splatter2017}, SPARSim\cite{SPARSim2019}, scGAN\cite{marouf2020realistic}, scVI\cite{Lopez2018}, SymSim\cite{SymSim2019}, have been designed to generate synthetic scRNA-seq datasets. These simulators employ various methodologies, including probability models\cite{WeiVivianLi2019,Risso2018,Splatter2017,SPARSim2019,Korthauer2016,scDesign2}, kinetic models\cite{SymSim2019}, and deep neural network models\cite{Lopez2018,marouf2020realistic}.  scDesign2, for instance, is a transparent simulator capable of producing high-fidelity synthetic data with gene correlations\cite{scDesign2}. It mimics the experimental scRNA-seq workflow steps, including cell isolation, RNA capture, library preparation, sequencing, and data analysis. 

Deep learning techniques have been extensively applied in biology and medicine research\cite{webb2018deep,jin2021application,ching2018opportunities}. In the realm of scRNA-seq, generative adversarial networks (GANs)\cite{goodfellow2020generative} have been employed for data synthesis\cite{wang2024generating}. For instance, the conditional single-cell GAN (cscGAN) was developed to learn nonlinear gene-gene dependencies in various cell type samples, generating class-specific cellular scRNA-seq\cite{marouf2020realistic}. While GANs have been successful, their training process instability and susceptibility to model collapse pose challenges. Alternatively, Variational Auto-Encoder (VAE)\cite{goodfellow2020generative} provides interpretability but may not guarantee high-quality scRNA-seq sample generation\cite{svensson2020interpretable}. 

Denoising Diffusion Probabilistic Models (DDPM) present a compelling alternative, capable of fitting complex and nonlinear distributions\cite{ho2020denoising}. Previous work attests to DDPM outperforms GAN in generative tasks\cite{dhariwal2021diffusion}. While commonly applied in computer vision and natural language processing, DDPM's success in bioinformatics applications\cite{guo2023diffusion}, such as generating RNA tertiary structures\cite{bafna2023diffrnafold}, brain MRI scans\cite{Kidder2023.08.18.553859}, and biological sequences design\cite{avdeyev2023dirichlet}, highlights its versatility. Previous works using diffusion models for generative tasks mostly use the UNet\cite{ronneberger2015u} neural network architecture as the skeleton\cite{wang2023single}. scRDiT pioneered Diffusion Transformers (DiTs)\cite{peebles2023scalable} , a method based on Vision Transformer (ViT)\cite{dosovitskiy2020image} that has been used in the field of computer vision, as the neural network architecture for diffusion models to synthesize scRNA-seq samples. The DiTs-based DDPM shows better performance than that of the UNet-based in this task. What's more, we try zero-negation, a special data preprocessing method, which comprehensively considers the characteristics of diffusion model and scRNA-seq data, and further improves the quality of the synthesized samples.

Despite its efficiency, DDPM faces challenges in sample generation speed compared to VAE and GAN. Denoising Diffusion Implicit Models (DDIM), a more efficient iterative variant with the same training process as DDPM\cite{song2020denoising}, addresses this limitation.

This paper introduces scRNA-seq Diffusion Transformer (scRDiT), a DiTs-based approach that leverages DDIM to generate high-quality scRNA-seq samples, striking a balance between efficiency and reliability. Specialized models for different cell types exhibit similarity to real samples, and DDIM accelerates the sampling process without unacceptable degradation in performance, achieving speedups of 10$\times$ to 20$\times$. Dimensionality reduction analysis and comparison with real data distributions using various metrics showcase the method's commendable results and improved synthesis efficiency.

\section{Materials and methods}
\label{sec:method}

In this section, we first briefly introduce the fundamentals of denoising diffusion probabilistic models (DDPMs) and then elucidate the architecture of our denoising network. Next, we present a general training method for this network, which can be used on different datasets. Finally, we provide a concise overview of the sampling accelerating method from denoising diffusion implicit models (DDIMs).

\subsection{Data prepare}
\label{sec:datasets}

The numerical framework of scRDiT consists of two steps: (1) model training and (2) synthetic data generation. scRDiT trains a diffusion model based on a real scRNA-seq dataset in the model training step. If the dataset contains more than one cell type, scRDiT divides the dataset into subsets, one per cell type, and fits a cell type-specific model to each subset.

\begin{figure}[htbp]
    \centering
    \includegraphics[width=1\textwidth]{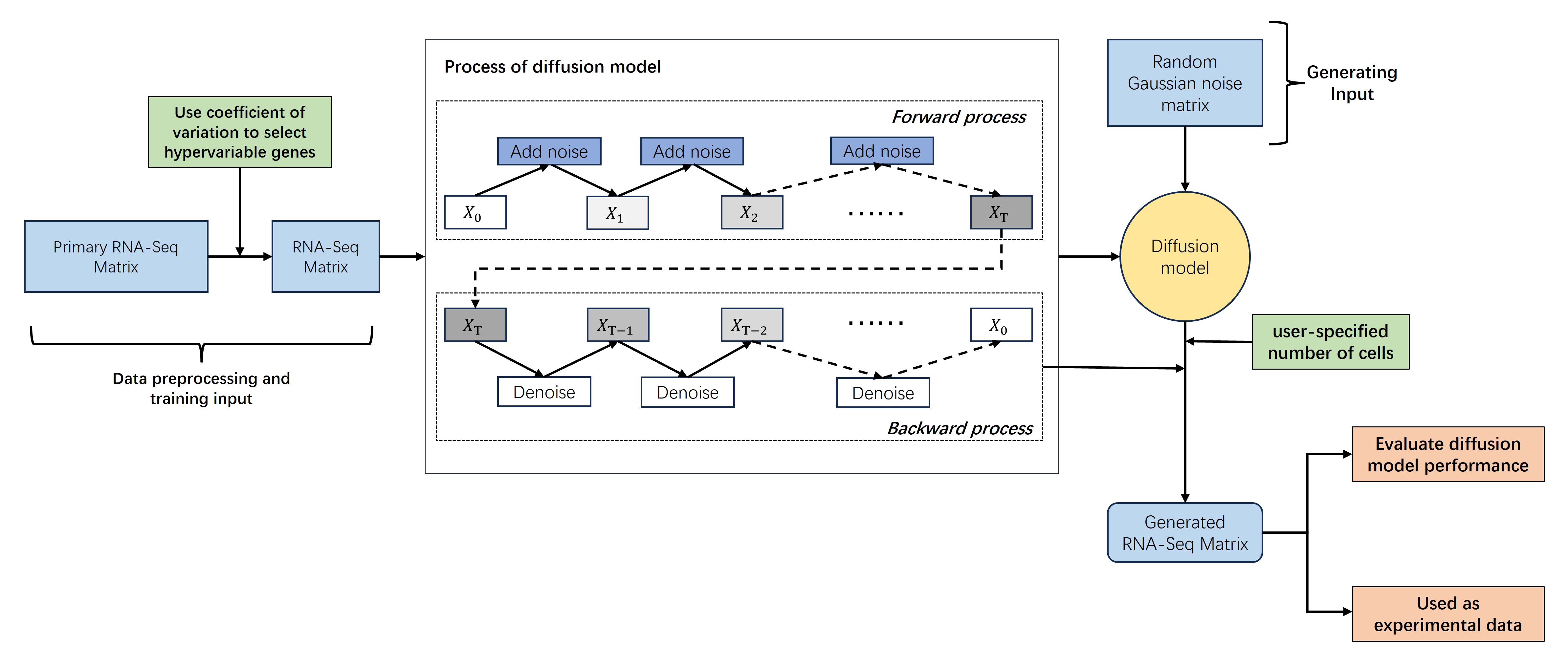}
    \caption{This figure shows how scRDiT process on scRNA-seq data.After data preprocessing, the real scRNA-seq data were fed into the model for training to fit the real data. After training, the user specifies the batch size of simulated data to be synthesized, and the model randomly generates Gaussian noise and reverts it to simulated scRNA-seq data.The generated simulated data can be used for downstream tasks or to test the reliability of the data.}
    \label{fig:data_process}
\end{figure}

Before model training, scRDiT screens out hypervariable genes through the coefficient of variation (cv) of gene expression levels. Consider gene $i$ and the vector $\vec{y}_i$ of its expression values in each cell, the coefficient of variation $cv_i$ for gene $i$ is\cite{reed2002use}:
\begin{equation}
cv_i = \frac{\mathrm{sd}(\vec{y}_i)}{\mathrm{mean}(\vec{y}_i)},
\end{equation}
where $\mathrm{sd}(\vec{y}_i)$ represents the standard deviation of the components of vector $\vec{y}_i$, and $\mathrm{mean}(\vec{y}_i)$ represents their mean. We arrange the coefficient of variation of all genes from large to small and select the largest $2000$ genes as the gene set of model fitting. Filtering out the genes containing less information can reduce the dimensionality of the data and facilitate downstream analysis.


The denoising process of DDPMs has some limitations when it comes to generating specific discrete values. In particular, the generation of zero is crucial in scRNA-seq samples. When fitting directly to the raw data, it is possible for genes that we expect to be expressed as zero to fluctuate in a vary small range around zero without any integer zero appearing in the synthetic samples. The issue can seriously impact the quality of the generated data. 

To resolve this problem, we adopt a zero-negation, which change zero in the original data to a negative number $n$ during training, so as to strengthen the characteristics of the data. (Practically, we take $n = -10$.) Finally, we truncate the data that were negative in the generated scRNA-seq sample to zero. Experiment results in section\ref{sec:comparison} show that this method significantly improves the reliability of the synthesized samples.


\subsection{Diffusion models}
\label{sec:diffusion-models}
The diffusion models are probabilistic models that generate target data samples from Gaussian noise by iterative denoising process. The diffusion model consists of two steps, the forward and reverse processes. The forward process is also called the diffusion process.  Either the forward or backward process is a parameterized Markov chain, and the backward process can be used to generate data samples. The complete process of the diffusion model is shown at Figure \ref{fig:data_process}.


Consider a real scRNA-seq sample $\vec{x}_0\sim{q(\vec{x}_0)}$, where $q(\cdot)$ represents the distribution. The forward process iteratively adds Gaussian noise to the real sample. For each step $t$ of the forward process, we add noise to the previous step $\vec{x}_{t-1}$ and get a new dataset $\vec{x}_t$. The dataset $\vec{x}_{t}$ can be represented as a weighted summation of $\vec{x}_{t-1}$ and a Gaussian noise $\vec{z}\sim{\mathcal{N}(\vec{0}, \vec{I})}$. Thus, the relationship between $\vec{x}_t$ and $\vec{x}_{t-1}$ is written as:
\begin{equation}
\label{eq:one-step-add-noise}
\vec{x}_t = \sqrt{\alpha_t} \vec{x}_{t-1}+\sqrt{1-\alpha_t} \vec{z}\quad,\quad \mathrm{where}\quad \vec{z}\sim{\mathcal{N}(\vec{0}, \vec{I})}.
\end{equation}
The parameter $\alpha_t$ is defined as:
\begin{equation}
    \alpha_t = 1-\beta_t
\end{equation}
where $\beta_t$ is a predefined hyperparameter. The parameters $\{\beta_t\}_{1\leq t\leq T}$ is an increase series so that 
\begin{equation*}
0 <  \beta_1 < \beta_2 < \cdots < \beta_T < 1.
\end{equation*}
From Eq.\eqref{eq:one-step-add-noise}, $\vec{x}_t$ only depends on $\vec{x}_{t-1}$ at the previous time step. This process can be considered a Markov process:
\begin{equation}
\begin{aligned}
    q(\vec{x}_{T} | \vec{x}_0) &= \prod_{t=1}^T q(\vec{x}_t | \vec{x}_{t-1}),\\
    q(\vec{x}_t | \vec{x}_{t-1}) &= \mathcal{N}(\sqrt{1-\beta_t}\vec{x}_{t-1}, \beta_t\vec{I})
\end{aligned}
\end{equation}
This process brings $\vec{x}_t$ closer and closer to pure noise as $t$ increases. When $T\rightarrow\infty$, $\vec{x}_t$ becomes a Gaussian noise. The above iterative process yield the relationship between $\vec{x}_t$ and $\vec{x}_0$ as:
\begin{equation}
q(\vec{x}_t | \vec{x}_0) = \mathcal{N}(\sqrt{\bar{\alpha}_t} \vec{x}_0, (1 - \bar{\alpha}_t)\vec{I}),\
    \mathrm{where}\ \bar{\alpha}_t = \prod_{i=1}^t \alpha_i.
\end{equation}

The reverse process of the diffusion model is the denoising process. We can restore the original distribution of $\vec{x}_0$ from the complete standard Gaussian distribution $\vec{x}_T\sim{\mathcal{N}(\vec{0}, \vec{I})}$ if we can obtain the reversed distribution $q(\vec{x}_{t-1}|\vec{x}_t)$ step by step. 

To obtain the reversed distribution, we use neural network $p_\theta(\vec{x}_{t-1} | \vec{x}_t)$ to fit the reverse process $q(\vec{x}_{t-1} | \vec{x}_t)$. If $\vec{x}_0$ is known, the reverse process can be obtained through the Bayesian formula
\begin{equation}
q(\vec{x}_{t-1} | \vec{x}_t; \vec{x}_0) = \mathcal{N}(\tilde{\mu}_t(\vec{x}_t, \vec{x}_0),\tilde{\beta}_t\vec{I}),
\end{equation}
where $q(\vec{x}_{t-1} | \vec{x}_t; \vec{x}_0)$ means the conditional distribution $p(\vec{x}_{t-1} | \vec{x}_t)$ given the sample data $\vec{x}_0$, and 
$$
\tilde{\mu}_t = \dfrac{\sqrt{\bar{\alpha}_t}\beta_t}{1-\bar{\alpha}_t}\vec{x}_0 + \dfrac{\sqrt{\alpha_t} (1 - \bar{\alpha}_t)}{1 - \bar{\alpha}_t}\vec{x}_t,\ \mathrm{and}\ \tilde{\beta}_t = \dfrac{1 - \bar{\alpha}_{t-1}}{1 - \bar{\alpha}_t} \beta_t. 
$$
Then, we can deduce  
\begin{equation}
    p_{\theta}(\vec{x}_{t-1} | \vec{x}_t) = \mathcal{N}(\mu_{\theta}(\vec{x}_t, t),\sigma_{\theta}(\vec{x}_t,t)),
\end{equation}
where $\mu_\theta(\vec{x}_t, t)$ represents the predicted expectation expressed as
$$
\mu_\theta(\vec{x}_t, t) = \dfrac{1}{\sqrt{\alpha_t}}\left(\vec{x}_t - \dfrac{1 - \alpha_t}{\sqrt{1 - \bar{\alpha}_t}}\vec{\epsilon}_\theta(\vec{x}_t, t)\right),\ \mathrm{and}\ \sigma_\theta(\vec{x}_t, t) = \sigma_t^2 \vec{I},
$$
with $\epsilon_\theta(\vec{x}_t, t)$ the predicted noise perturabtion. 

If we ignore the variance of the distribution above and fit the mean with the neural network, we obtain the sampling process of the diffusion model as\cite{ho2020denoising}:
\begin{equation}
\label{equ:ddpm_sample}
\vec{x}_{t-1} = \frac{1}{\sqrt{\alpha_t}} \left(\vec{x}_t - \frac{1-\alpha_t}{\sqrt{1-\bar{\alpha}_t}}\vec{\epsilon}_{\theta}(\vec{x}_t,t)\right)+\sigma_t \vec{z},\
    \mathrm{where}\ \vec{z}\sim{\mathcal{N}(\vec{0}, \vec{I})}.
\end{equation}
Practically, we usually take $\sigma_t^2 = \beta_t$. Since $t$ and $\vec{x}_t$ are known, we only need to fit $\vec{\epsilon}_{\theta}(\vec{x}_t,t)$ with neural network.

\subsection{Neural network architecture}
\label{sec:unet}
We design a neural network architecture as the noise predictor with reference to the DiTs\cite{peebles2023scalable}. A scRNA-seq sample with Gaussian noise and a timestep $t$ are fed into the model to obtain the predicted noise. For a noised sample with $n$ genes as the input tensor of dimension $(n,1)$, "Patchify" layer converts the tensor into a sequence of $n / p$ tokens with convolutional layer, where $p$ is a hyperparameter of patch size. Each token is a vector of length $h$, where hidden size $h$ is also a hyperparameter that could be adjusted for different tasks. "Patchify" layer's output should be a tensor $P$ of dimension $(n/p, h)$, and timestep embed layer will embed $t$ into a tensor $T$ of the same dimension. Pass $T$ and $P$ into the scalable Transformer unit with $N\times$ DiT blocks\cite{peebles2023scalable} and then to the final layer. The final output of the model is the predicted noise for timestep $t$.


\begin{figure}[htbp]
    \centering
    \includegraphics[width=0.6\textwidth]{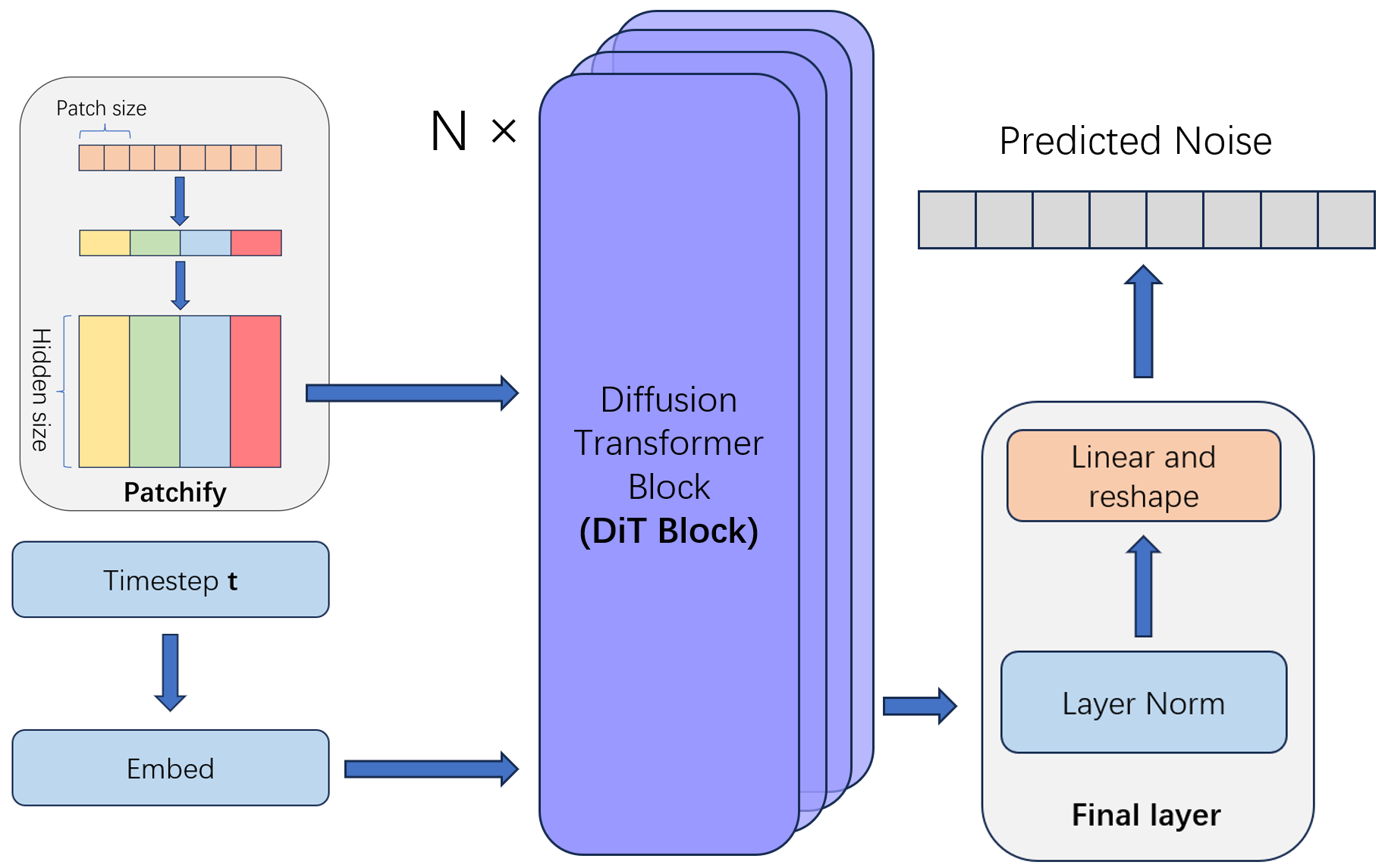}
    \caption{Illustration of the DiT architecture.}
    \label{fig:DiT}
\end{figure}

\subsection{Training}
Here, we show the training method of DiT. DiT requires multiple training epochs in order to achieve convergence. The number of required epochs is determined according to the dataset's characteristics. 

At the beginning of each epoch, we first draw a sample $\vec{x}_0\sim{q(\vec{x}_0)}$ from the dataset. Randomly pick a time step $t$ from $\{1, 2, \cdots, T\}$ following
$$t\sim\mathrm{Uniform}(\{1, \cdots, T\}).$$
Next, we add Gaussian noises of time step $t$ to the sample $\vec{x}_0$ as:
\begin{equation}
    \vec{x}_t = \sqrt{\overline{\alpha}_t}\vec{x}_0 + \sqrt{1-\overline{\alpha}_t}\vec{\epsilon}
\end{equation}
where $\vec{\epsilon}\sim{\mathcal N(\vec{0}, \vec{I})}$, $\alpha_t = 1 - \beta_t$, and $\overline{\alpha}_t = \prod_{i=1}^t \alpha_i$.
Then, we pass $\vec{x}_t$ and $t$ to DiT as parameters. DiT outputs the predicted noise $\vec{\epsilon}_{\theta}$ through the given $\vec{x}_t$ and $t$. Use $\vec{\epsilon}_{\theta}$ and the true Gaussian noise $\vec{\epsilon}$ to calculate the loss $\mathcal{L}$ function as\cite{ho2020denoising}:
\begin{equation}
\label{equ:ddpm_loss}
    \mathcal{L} = \mathbb{E}_{t,  \vec{x}_0, \vec{\epsilon}}\Big[\|\vec{\epsilon} - \vec{\epsilon}_{\theta}(\sqrt{\overline{\alpha}_t} \vec{x}_0 + \sqrt{1-\overline{\alpha}_t}\vec{\epsilon}, t)\|^2\Big].
\end{equation}
At the end of each epoch, apply the backpropagation to calculate the gradient and update the weights of DiT. Repeat the steps above until the model converges.

\subsection{Sampling acceleration}
\label{sec:sampling acceleration}

It is well-known that the denoising diffusion probabilistic model (DDPM) has a major drawback compared to most other generative models, which is its slow sampling speed. This is because DDPM typically requires 1000 denoising operations for each sampling process, resulting in a high time cost when using DDPM to generate sample data on low-performance machines. In the field of image generation, a common solution to this problem is to use the denoising diffusion implicit model (DDIM), a much faster sampling method. We applied this method to the task of generating scRNA-seq data and obtained satisfactory results.

The DDIM sampling method removes the Markov assumption of DDPM by enabling us to sample using a subset of time steps $\{1, 2, \cdots, T\}$. This means that we can directly predict the outcome after several denoising steps following a particular denoising step. For instance, if the model starts denoising at $T=1000$, a DDIM-based model can predict the outcome after 10 denoising steps, including  $t=990, 980, \cdots,10,1$. This results in a ten-fold acceleration of the sampling process.

To be able to accelerate the sampling process, we need to change the sampling method of DDPM to:
\begin{equation}
\label{equ:ddim_sample}
\vec{x}_{t-1} = \sqrt{\bar{\alpha}_{t-1}}\bigg(\frac{\vec{x}_t - \sqrt{1-\bar{\alpha}_t}\vec{\epsilon}_{\theta}}{\sqrt{\bar{\alpha}_t}}\bigg) + \sqrt{1 - \bar{\alpha}_{t-1} - \sigma_{t}^2}\vec{\epsilon}_{\theta} + \sigma_t\vec{\epsilon}_t,
\end{equation}
\begin{equation}
\label{equ:sigma_t}
\sigma_t = \eta\sqrt{\frac{1 - \bar{\alpha}_{t-1}}{1 - \bar{\alpha}_t}}\sqrt{1 - \frac{\bar{\alpha}_t}{\bar{\alpha}_{t-1}}},\quad\eta \in [0,1],
\end{equation}
where $\vec{\epsilon}_t\sim{\mathcal{N}(\vec{0}, \vec{I})}$ is independent of $\vec{x}_t$, and $\vec{\epsilon}_{\theta}$ is the predicted noise of $\vec{x}_t$ and time step $t$.\cite{song2020denoising} Here, $\eta$ is a hyper-parameter to adjust the acceleration efficiency. Particularly, when $\eta=1$, DDIM is equivalent to DDPM. 

To apply DDIM to accelerate the sampling process, we select a sub-sequence $\tau$ of sampling steps, where:
$$
\tau = \{\tau_1, \tau_2, \cdots, \tau_{i}, \cdots\} \subset \{1, \cdots, T\},
$$
and formulate the sampling method (\ref{equ:ddim_sample}) and (\ref{equ:sigma_t}) as:
\begin{equation}
    \vec{x}_{\tau_{i-1}} = \sqrt{\bar{\alpha}_{\tau_{i-1}}}\bigg(
    \frac
    {\vec{x}_{\tau_i} - \sqrt{1-\bar{\alpha}_{\tau_i}}\vec{\epsilon}_{\theta}}
    {\sqrt{\bar{\alpha}_{\tau_i}}}
    \bigg) +
    \sqrt{1 - \bar{\alpha}_{\tau_{i-1}} - \sigma_{\tau_i}^2}\vec{\epsilon}_{\theta}
    + \sigma_{\tau_i} \vec{\epsilon}_{\tau_i},
\end{equation}
\begin{equation}
    \sigma_{\tau_i} = \eta\sqrt{\frac
    {1 - \bar{\alpha}_{\tau_{i-1}}}
    {1 - \bar{\alpha}_{\tau_i}}
    }
    \sqrt{1 - \frac
    {\bar{\alpha}_{\tau_i}}
    {\bar{\alpha}_{\tau_{i-1}}}
    },\quad
    \eta \in [0,1].
\end{equation}

It's important to keep in mind that using the loss function $\mathcal{L}$ in the form of \eqref{equ:ddpm_loss} during the training of neural networks, such as DiTs, as noise predictors, can lead to direct use of the training results in the sampling process of DDIM. This means that we can use the same neural network model, which was trained for DDPM, to speed up the sampling process by DDIM without having to retrain it\cite{song2020denoising}.

\section{Results}
\label{sec:results}

In implementing the DiT model structure as a noise predictor, we trained multiple models to generate scRNA-seq samples of the datasets. In addition, we tested different neural networks and conducted many comparisons to demonstrate the superiority of DiT architecture in this task. Using zero-negation preprocessed data as the training set (section \ref{sec:datasets}) , we enhance the adaptation of the diffusion model to scRNA-seq data. Two sampling methods, Denoising Diffusion Probabilistic Models (DDPM) and Denoising Diffusion Implicit Models (DDIM), were employed, and comprehensive comparative experiments were conducted. Notably, DDIM demonstrated effective acceleration of the sampling process. Furthermore, we tested various acceleration rates and compared the generated sample results from the models. The findings indicate that DDIM’s sampling method is faster than that of DDPM and maintains high-quality samples. Importantly, the control over the model's performance loss remains within an acceptable range.

\subsection{Training results}

\begin{figure}[!htb]
    \centering
    \includegraphics[width=1\textwidth]{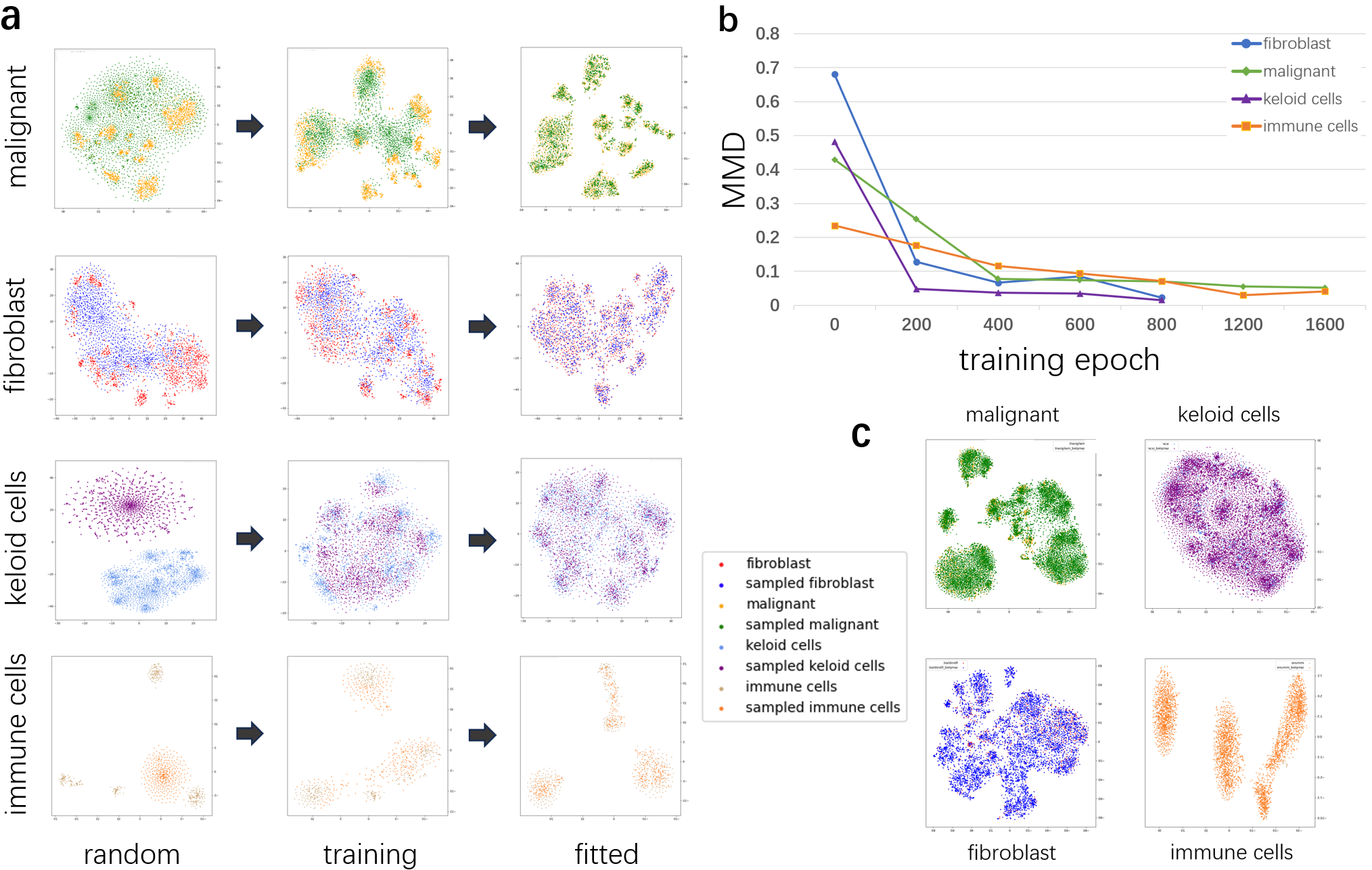}
    \caption{scRDiT generating scRNA-seq data. \textbf{a.} The fitting process of the model. t-SNE plots of real data and generated scRNA-seq samples from beginning of training to the final performance. \textbf{b.} Dependence of MMD with the increase of training epochs.
 \textbf{c.} t-SNE plots of virtual scRNA-seq samples and real data.}
    \label{fig:training}
\end{figure}

We applied the neural network architecture in section \ref{sec:unet} to train two models for generating fibroblast and malignant scRNA-seq data, respectively. We employed scRDiT to assess its effectiveness on a tumor cell dataset (GSE103322), specifically consisting of 5902 cells derived from 19 patients with oral cavity head and neck squamous cell carcinoma\cite{puram2017single}. The dataset was partitioned into two subsets: 2215 malignant tumor cells and 1422 fibroblasts. Usually, different epochs are required for different datasets due to the variation in data features. For the models trained with different epochs, we generated a set of virtual data through DDPM's sampling method and compared the similarity between the generated data and real data. Various measurements were used to check the sample quality, including the KL divergence, Wasserstein distance, and maximum mean discrepancy (MMD). The results are shown in Table\ref{tab:training results} and Figure\ref{fig:training}-b. 

We use t-SNE dimensionality reduction\cite{van2008visualizing} to reduce real and generated data samples to two dimensions and make scatter plots. The fitting process for different kinds of cells are shown in Figure \ref{fig:training}-a. The result shows that as the number of training epochs increases, the data distribution of the generated samples becomes closer to the real samples.

Moreover, we used two data sets for model training to validate the method's generality. The first comprised 407 immune cells from the GSE103322 dataset\cite{puram2017single}, while the second incorporated a dataset (GSE181297) of 1936 keloid cells\cite{Shim2022}. The results are also shown in Table\ref{tab:training results} and Figure \ref{fig:training}-b.

The trained model is robust in generating sampling data. We generated 5k scRNA-seq samples for each category with the best-performing models. The results of t-SNE dimensionality reduction for all samples are shown in Figure \ref{fig:training}-c. The real data and the generated samples show good consistency, which implies the robustness and stability of our method.

\begin{table}[htbp]
\renewcommand\arraystretch{1.2}
\label{tab:training results}
\centering
\caption{Sample qualities of different cell types. The models are trained with different training epochs. The qualities are measured with different methods: KL divergence, Wasserstein distance, and MMD. (Bolded numbers for the best results.)}

\begin{tabular}{|c|cccc|cccc|}
\hline
single-cell RNA-seq type & \multicolumn{4}{c|}{fibroblast}        & \multicolumn{4}{c|}{malignant} \\ \hline
epoch                    & 200   & 400   & 600   & 800            & 400    & 800   & 1200  & 1600  \\ \hline
KL Divergnece            & 4.570 & 4.096 & 4.016 & 3.971          & 3.798  & 3.117 & 3.648 & 3.179 \\ \cline{1-1}
Wasserstein Distance     & 0.052 & 0.047 & 0.052 & 0.026          & 0.061  & 0.045 & 0.023 & 0.007 \\ \cline{1-1}
MMD                      & 0.128 & 0.066 & 0.084 & \textbf{0.022} & 0.077  & 0.070 & 0.055 & \textbf{0.046} \\ \hline
\end{tabular}

\begin{tabular}{|c|cccc|cccc|}
\hline
single-cell RNA-seq type & \multicolumn{4}{c|}{keloid cells}      & \multicolumn{4}{c|}{immune cells}      \\ \hline
epoch                    & 200   & 400   & 600   & 800            & 400   & 800   & 1200           & 1600  \\ \hline
KL Divergnece            & 0.149 & 0.145 & 0.143 & 0.151          & 6.455 & 7.208 & 7.405          & 7.357 \\ \cline{1-1}
Wasserstein Distance     & 0.011 & 0.008 & 0.008 & 0.005          & 0.120 & 0.023 & 0.015          & 0.043 \\ \cline{1-1}
MMD                      & 0.047 & 0.036 & 0.034 & \textbf{0.015} & 0.116 & 0.071 & \textbf{0.029} & 0.041 \\ \hline
\end{tabular}


\end{table}

\subsection{Comparison}
\label{sec:comparison}

The previous work mostly use UNet\cite{ronneberger2015u} architecture as the noise predictor\cite{wang2023single} for DDPMs. In order to verify the superiority of DiT\cite{peebles2023scalable} network architecture, we tested different neural network architectures to synthesize scRNA-seq samples. The performance comparison in Table\ref{tab:network-comparison} shows that DiT is able to generate the most plausible samples.

\begin{figure}
    \centering
    \includegraphics[width=1\textwidth]{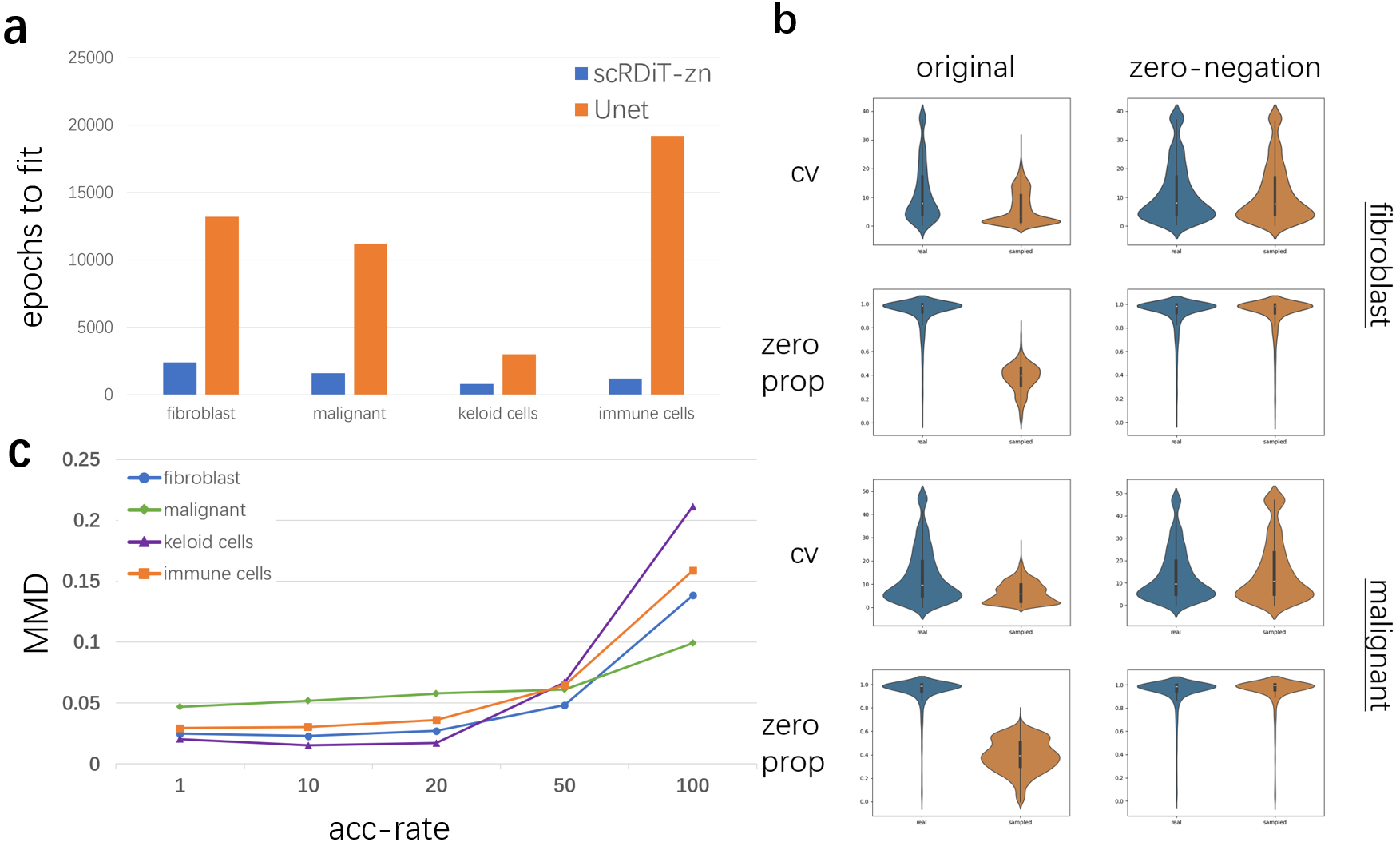}
    \caption{Comparison with scRDiT and UNet. \textbf{a.} Number of epochs to fit real data based on DiT and UNet. \textbf{b.} Violin plots for the coefficient of variation (cv) and zero proportion (zero prop) of scRNA-seq data for the two methods. \textbf{c.} Effect of different acceleration rates on the quality of the synthesized samples.}
    \label{fig:comparison}
\end{figure}

UNet models require much more training epochs  to fit the real data than DiT models (Figure\ref{fig:comparison}-a) . It is worth mentioning that each training epoch and each sampling process of UNet models is much slower than that of DiT models.

Most methods that apply deep learning to generate scRNA-seq samples obtain results that have problems in the simulation of coefficient of variation and zero proportion. Preprocessing the training data using zero-negation, introduced in section\ref{sec:datasets}, can effectively solve this problem. The results of this method on different type of samples are shown in Figure\ref{fig:comparison}-b.

\begin{table}[htbp]
\renewcommand\arraystretch{1.2}
\centering
\label{tab:network-comparison}
\caption{Performance of different neural networks on the same scRNA-seq datasets. scRDiT-zn using special preprocessing method, zero-negation, on the training set and obtained the best results on each type of samples.}
\begin{tabular}{lllll}
\hline
                   & \multicolumn{4}{c}{MMD}                                                                                                              \\ \hline
Model              & \multicolumn{1}{c}{fibroblast} & \multicolumn{1}{c}{malignant} & \multicolumn{1}{c}{keloid cells} & \multicolumn{1}{c}{immune cells} \\ \hline
UNet               & 0.039                          & 0.066                         & 0.036                            & 0.054                            \\
scRDiT             & 0.065                          & 0.053                         & 0.028                            & 0.061                            \\
\textbf{scRDiT-zn} & 0.022                          & 0.046                         & 0.015                            & 0.030                            \\ \hline
\end{tabular}
\end{table}

\subsection{Sampling acceleration}

To delve deeper into the acceleration method outlined in Section \ref{sec:sampling acceleration}, we implemented the DDIM sampling approach in the sampling processes. We specifically opted for various time-step sub-sequences to attain different acceleration rates. The qualities of the sampled data are detailed in Table \ref{tab:acc}. Additionally, the variation of the models' performance (measured with MMD between real data and sampled data) under different acceleration rates is shown in Figure\ref{fig:comparison}-c. The outcomes indicate that the quality of the sampled data remains consistent, while the sampling speeds experience a noticeble increase. It is possible to greatly speed up the generation of scRNA-seq samples using the DDIM sampling method. However, increasing acceleration beyond a certain point can have a negative impact on the quality of generated samples. In our experiments, we found that an acceleration rate of 10 to 20 times strikes a balance between time consumption and sample quality.

For instance, generating 2k scRNA-seq samples using an NVIDIA RTX4060 GPU device typically takes about 50 minutes without acceleration, while it only takes less than 5 minutes with 10$\times$ acceleration. This means that by using the accelerated sampling method, we can generate many samples at an acceptable time cost on a low-performance device while ensuring time cost on a low-performance device while ensuring high sample quality. This partly overcomes the problem that the diffusion model is slower to do generative tasks than other generative models such as GAN. Furthermore, we found that using non-equidistant time steps in sub-sequences for sampling can further accelerate the process while maintaining high sample quality.


\begin{table}[htbp]
\renewcommand\arraystretch{1.2}
\centering
\label{tab:acc}
\caption{The similarity between the generated scRNA-seq samples and the real samples under different acceleration rates.}

\begin{tabular}{|c|ccccc|ccccc|}
\hline
scRNA-seq type & \multicolumn{5}{c|}{fibroblast}       & \multicolumn{5}{c|}{malignant}         \\ \hline
Acc-Rate                 & 1     & 10    & 20    & 50    & 100   & 1     & 10    & 20    & 50    & 100   \\ \hline
KL Divergnece            & 4.057 & 3.971 & 3.988 & 3.835 & 3.294 & 3.179 & 3.137 & 3.117 & 3.052 & 2.960 \\ \cline{1-1}
Wasserstein Distance     & 0.037 & 0.026 & 0.026 & 0.028 & 0.072 & 0.007 & 0.007 & 0.007 & 0.012 & 0.029 \\ \cline{1-1}
MMD                      & 0.024 & 0.022 & 0.027 & 0.048 & 0.138 & 0.046 & 0.051 & 0.057 & 0.061 & 0.099 \\ 
\hline
\hline
scRNA-seq type & \multicolumn{5}{c|}{keloid cells}             & \multicolumn{5}{c|}{immune cells}           \\ \hline
Acc-Rate                 & 1     & 10    & 20    & 50    & 100   & 1     & 10    & 20    & 50    & 100   \\ \hline
KL Divergnece            & 0.157 & 0.151 & 0.149 & 0.132 & 0.127 & 7.405 & 7.097 & 7.230 & 6.995 & 6.629 \\ \cline{1-1}
Wasserstein Distance     & 0.006 & 0.005 & 0.006 & 0.012 & 0.020 & 0.015 & 0.012 & 0.015 & 0.029 & 0.039 \\ \cline{1-1}
MMD                      & 0.020 & 0.015 & 0.017 & 0.066 & 0.211 & 0.029 & 0.030 & 0.036 & 0.064 & 0.158 \\ \hline
\end{tabular}

\end{table}

\section{Summary}
\label{sec:summary}

In this paper, we propose a novel approach for generating scRNA-seq samples, utilizing the diffusion model. This method provides a viable solution to the challenge of limited available scRNA-seq samples. Employing the DiTs structure as a reference, we trained neural networks to serve as noise predictors, and achieved better results than several other neural networks. Our model effectively reconstructs simulated scRNA-seq samples from Gaussian noise. To enhance synthesis efficiency and address the prolonged sampling processes of DDPM, we leverage DDIM. In summary, our method demonstrates superior performance in generating synthetic scRNA-seq samples akin to the datasets we utilized, exhibiting excellent stability and speed.

\section*{Acknowledgments}
DL is supported by the Tianjin Natural Science Foundation of China (20JCYBJC00500), 
the Science \& Technology Development Fund of Tianjin Education
Commission for Higher Education (2018KJ217). JL is supported by the National Natural Science Foundation of China (12331018)

\bibliographystyle{unsrt}  
\bibliography{references}

\end{document}